\documentclass[runningheads]{llncs}

\usepackage{eccv}

\usepackage{eccvabbrv}

\usepackage{graphicx}
\usepackage{booktabs}
\usepackage{xcolor}
\usepackage{pifont}
\definecolor{cmarkgreen}{RGB}{20,120,60}
\newcommand{\cmark}{\textcolor{cmarkgreen}{\ding{51}}}
\newcommand{\xmark}{\textcolor{gray!55}{\ding{55}}}

\usepackage[accsupp]{axessibility}

\usepackage{hyperref}

\usepackage{orcidlink}

\begin{document}

% ---------------------------------------------------------------
\title{VLCP: Vision Language Control Policy\\
Closed-Loop Code Replanning for Robot Manipulation}

\titlerunning{VLCP: Vision Language Control Policy}

\author{Dhia Naouali\inst{1} \and
Minghan Wu\inst{2} \and Claudia Wong\inst{3} \and
Abhinav Puthran\inst{4} \and Omar G. Younis\inst{5}\inst{6}}
\authorrunning{D. Naouali et al.}
\institute{
$^{1}$University of Monastir,\ 
$^{2}$St. Mildred's-Lightbourn School,\ 
$^{3}$Cornell University,\ 
$^{4}$Carnegie Mellon University,\ 
$^{5}$Silverstream AI,\
$^{6}$Mila -- Quebec AI Institute
}
\footnotetext{Corresponding author: \email{dhia12naouali@gmail.com}}
\maketitle

\begin{abstract}
Turning a frontier vision-language model into a robot policy usually means
fine-tuning it to emit an action representation it never saw in pretraining,
which throws away much of the reasoning that made the model worth reaching for.
We go the other way and keep the VLM \emph{frozen}. It writes the policy as a
short Python \texttt{control} function, with no demonstrations and no
fine-tuning. Writing that code once is open-loop, though. Existing closed-loop
methods react at the wrong level: they retry a fixed policy or pick a different
subtask, but never rewrite the code that failed. \textbf{VLCP} closes the loop
where the failure actually lives, on the control code, within a single episode.
Every $K$ steps the VLM re-observes the scene from multi-view RGB, proprioceptive
state, and a state delta, then rewrites the control function from what it just
saw, so a failure is caught before it compounds.

We evaluate on a 57-task MuJoCo/RoboVerse sweep. This training-free policy
reaches $35.1\%$ pooled success, against $3.5\%$ for the identical system
queried once per episode. That tenfold gap holds with non-overlapping
confidence intervals in every scene family. The gain traces to a $27.3\%$ within-episode recovery
rate on failed grasps: a miss an open-loop controller would carry to the end of
the episode gets re-observed and fixed at the next replan. And the loop stays
cheap. A median $84\%$ of input tokens hit cache, an episode needs only about
$10$ compact queries, and control blocks written during any replan persist to a
cross-episode skill library reused in later prompts.

\keywords{robot manipulation \and code-as-policy \and vision-language models
\and closed-loop control \and replanning}
\end{abstract}

\begin{figure}[t]
\centering
\includegraphics[width=\linewidth]{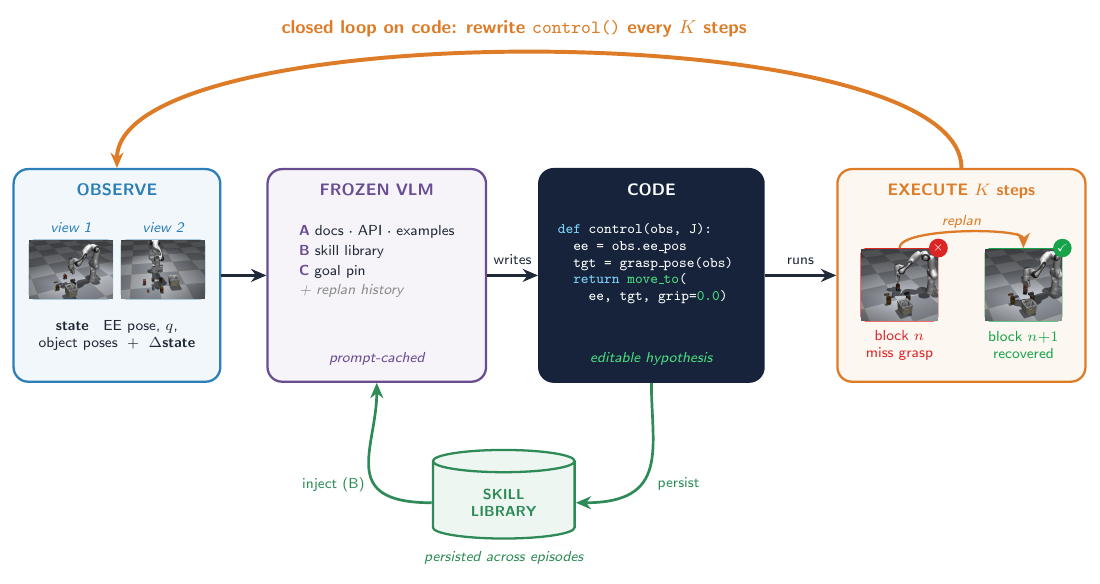}
\caption{\textbf{VLCP closes the loop at the level of the control code.}
A frozen VLM observes synchronized multi-view RGB, numeric state, and an
explicit state delta (\textsc{observe}), then \emph{writes} a short-horizon
Python \texttt{control(obs,\,J)} function (\textsc{code}) that runs open-loop
for $K$ steps (\textsc{execute}). At each block boundary the VLM is re-queried
and rewrites the function from what it just saw, so a failure is caught and
corrected \emph{within the same episode} (block $n{+}1$). The prompt is
structured for cache reuse (A~docs/API/examples, B~skill library, C~goal pin),
and helper modules written during any replan are saved to a cross-episode skill
library and re-injected into later prompts.}
\label{fig:method}
\end{figure}

\section{Introduction}
\label{sec:intro}
The dominant way to put multimodal LLMs to work in robotics has been the
vision-language-action model (VLA), which adapts a pretrained vision-language
backbone to emit low-level robot actions~\cite{openvla, pi0}. That adaptation is
not free. The model has to be fine-tuned on large demonstration datasets to map
observations onto an action representation it never saw during pretraining, and
every step away from the pretraining distribution risks throwing away the
reasoning ability that motivated reaching for an LLM in the first place.
VLA-0~\cite{vla0} put a number on this cost: representing actions as \emph{plain
text} instead of specialized tokens shrinks the distribution mismatch and yields
a markedly stronger policy. The lesson is general. The less an adaptation drags a
pretrained model away from what it is already fluent in, the more of its
capability survives.

So what are today's frontier LLMs \emph{most} fluent in? Code. Rather than teach
the model a new action language, we let it control the robot in a language
it already commands: the policy is a short Python \texttt{control} function that
the model \emph{writes}. This stays inside the pretraining distribution, keeps
essentially all of the model's coding and reasoning ability, and needs \emph{no}
fine-tuning. It is the promise of code as an action
representation~\cite{liang2023cap, singh2023progprompt}.

Writing code once, before the arm moves, is blind to everything after execution
begins. A gripper that fails to grasp and proceeds as though it had carries that
failure silently to the end of the episode. To close the loop, manipulation needs
two things at once: control code expressive enough to inspect and revise, and
observation close enough to catch failures as they happen. Current methods give
one or the other (Table~\ref{tab:related}). \textbf{(1) Code-generation
methods}~\cite{liang2023cap, singh2023progprompt, ellmer} produce interpretable
programs but never re-query during execution, so a runtime failure is
unrecoverable. \textbf{(2) Monitoring and replanning
methods}~\cite{codeAsMonitor, innerMonologue, doremi} close the loop only at the
task or language level: they retry the same policy or reselect a subtask without
rewriting the control code. \textbf{(3) Vision-language-action
models}~\cite{vla0, openvla, pi0} react to observations but emit opaque motor
tokens, lean on large demonstration datasets, and expose no point for
mid-execution reasoning.

None of them close the loop on the control code itself \emph{within} a single
episode. The closest work, Act-Observe-Rewrite~\cite{aor2026}, also treats the
full controller as the unit of revision, but it rewrites \emph{between} trials:
an episode runs to completion under a fixed controller before the model sees what
happened. \textbf{VLCP} (\textbf{V}ision \textbf{L}anguage \textbf{C}ontrol
\textbf{P}olicy) re-queries a frozen VLM every $K$ steps from runtime multi-view
RGB, proprioceptive state, and a state delta, rewriting the control function from
what it observes while the episode is still unfolding. When the first function
misses a grasp, the next replan boundary sees the miss, rewrites the function
with a corrected approach, and grasps on the retry, all inside the same episode.

Our contributions:
\begin{enumerate}
    \item \textbf{VLCP}, a training-free closed-loop manipulation framework that
    re-queries a frozen VLM every $K$ steps to rewrite the control function from
    live multi-view RGB, proprioceptive state, and an explicit state delta.
    \item \textbf{A cross-episode skill-library mechanism.} Control code blocks
    written during any replan are extracted, saved to a shared on-disk
    directory, and re-injected verbatim into every subsequent replan's prompt,
    so a later \texttt{control} function can import a skill written earlier.
    \item \textbf{A controlled ablation that isolates the replanning loop.}
    Compared against its own open-loop degenerate case ($K{=}T$), which shares
    VLCP's simulator, API, task set, model, and episode budget and differs only
    in within-episode replanning, closed-loop VLCP lifts pooled success from
    $3.5\%$ to $35.1\%$: a tenfold gain, with non-overlapping confidence
    intervals in every scene family. That $35.1\%$ is a strong absolute number
    for a zero-demonstration method; a concurrent training-free code agent lands
    near $18$--$22\%$ on a harder, perturbed benchmark. But that agent also
    closes the loop on code, so it is the matched $K{=}T$ ablation, not the
    cross-benchmark number, that pins the gain to the loop itself
    (Section~\ref{sec:exp-main}).
    \item \textbf{A direct measurement of within-episode recovery.} Computed
    from the logged replan traces, $27.3\%$ of failed grasps are re-approached
    and lifted by a later replan, a transition the open-loop variant cannot make
    by construction. This is the mechanism behind the gain
    (Section~\ref{sec:exp-recovery}).
    \item \textbf{A feasibility analysis of the loop.} Prompt caching keeps
    within-episode replanning affordable: a median $84\%$ of input tokens are
    served from cache, at only $\sim$10 compact VLM calls per episode
    (Section~\ref{sec:exp-cost}).
\end{enumerate}

\section{Related Work}
\label{sec:related}

\begin{table}[t]
\centering
\caption{\textbf{Where prior families sit along the axes VLCP unifies.}
\emph{Code policy}: the executed controller is human-readable code.
\emph{Runtime re-obs.}: the method senses the scene \emph{during} the episode.
\emph{Edits code}: the revision mechanism rewrites the control program that
failed, instead of retrying it, swapping in a library entry, or escalating to a
planner. \emph{Within-ep.}: revision happens mid-episode, not only between
trials. \emph{Train-free}: no demonstration fine-tuning. Closing the loop on
the control code \emph{within} an episode (all five columns) is a very recent
frontier. To our knowledge, VLCP and the concurrent CaP-Agent0~\cite{capx2026}
are among the few methods that occupy it, and of the two it is VLCP that
contributes the controlled ablation isolating the loop
(Section~\ref{sec:exp-main}) and a direct measurement of within-episode recovery
(Section~\ref{sec:exp-recovery}).}
\label{tab:related}
\setlength{\tabcolsep}{5pt}
\footnotesize
\begin{tabular}{lccccc}
\toprule
Method & Code & Runtime & Edits & Within- & Train- \\
       & policy & re-obs. & code & ep. & free \\
\midrule
CaP / ProgPrompt / ELLMER~\cite{liang2023cap,singh2023progprompt,ellmer} & \cmark & \xmark & \xmark & \xmark & \cmark \\
Robotic Programmer~\cite{roboticprogrammer}                              & \cmark & \xmark & \xmark & \xmark & \cmark \\
Code-as-Monitor / DoReMi~\cite{codeAsMonitor,doremi}                     & \cmark & \cmark & \xmark & \cmark & \cmark \\
Inner Monologue~\cite{innerMonologue}                                    & \xmark & \cmark & \xmark & \cmark & \cmark \\
VLA-0 / OpenVLA / $\pi_0$~\cite{vla0,openvla,pi0}                        & \xmark & \cmark & \xmark & \cmark & \xmark \\
Act-Observe-Rewrite~\cite{aor2026}                                       & \cmark & \cmark & \cmark & \xmark & \cmark \\
CaP-Agent0 (CaP-X, concurrent)~\cite{capx2026}                           & \cmark & \cmark & \cmark & \cmark & \cmark \\
\midrule
\textbf{VLCP (ours)}                                                     & \cmark & \cmark & \cmark & \cmark & \cmark \\
\bottomrule
\end{tabular}
\end{table}

\textbf{Code generation for robot control.}
Code-as-Policies~\cite{liang2023cap} established code as a viable action
representation by prompting a language model to compose perception and control
primitives into executable programs. ProgPrompt~\cite{singh2023progprompt} added
control flow, ELLMER~\cite{ellmer} added retrieval and contact-rich feedback
checks, and Robotic Programmer~\cite{roboticprogrammer} generates code from video
demonstrations. All of them query the model once, before execution, so any error
after the arm starts moving goes uncorrected. VLCP re-queries at a fixed, short
control horizon, so a deviation is caught before it compounds.

\textbf{Closed-loop monitoring and replanning.}
A second line keeps the policy fixed and adds failure detection.
Code-as-Monitor~\cite{codeAsMonitor} writes constraint-checking code that triggers
a retry, DoReMi~\cite{doremi} checks language-derived constraints visually and
falls back to a fixed recovery routine, and Inner Monologue~\cite{innerMonologue}
feeds success detection and scene descriptions back into a high-level planner.
None of them edit the control code that produced the failure. They retry it, swap
in a library entry, or escalate to a planner, and none of those repairs a
low-level behavior with no plan-level correlate. The closest work,
Act-Observe-Rewrite~\cite{aor2026}, does rewrite the full controller, but
\emph{between} trials: after a rollout, an LLM inspects key frames and outcomes and
writes a new controller for the next trial. VLCP rewrites \emph{within} an episode,
every $K$ steps, interrupting a failure mid-execution. Concurrent work,
CaP-Agent0~\cite{capx2026}, reaches the same design point from the benchmarking
side, which is independent evidence that within-episode code rewriting is the
right locus. VLCP's distinct contribution is the controlled $K{=}T$ ablation
isolating the loop (Section~\ref{sec:exp-main}) and a direct measurement of the
recovery it buys (Section~\ref{sec:exp-recovery}).

\textbf{Vision-language-action models.}
VLA-0~\cite{vla0}, OpenVLA~\cite{openvla}, and $\pi_0$~\cite{pi0} map observations
straight to actions. They are reactive, but their action representation, whether
specialized tokens~\cite{openvla,pi0} or plain text~\cite{vla0}, leaves no point to
inspect or correct the model's reasoning mid-task, and they depend on large
demonstration datasets the training-free methods above do not need.

\textbf{Skill library.} Outside robotics, Voyager~\cite{voyager} showed an LLM
agent can accumulate a library of self-verified skills over an open-ended
curriculum. VLCP borrows that accumulate-and-reuse idea at a finer grain: helper
modules written during any replan are saved to a shared on-disk library and
surfaced verbatim in later prompts, without the curated library, promotion
criteria, or curriculum Voyager maintains.

\section{Method}
\label{sec:method}

\subsection{Problem formulation}

We consider episodic manipulation tasks over a horizon $T$. At each step $t$ the
agent receives a multi-view RGB observation from two fixed cameras, a
proprioceptive state vector (end-effector pose, joint positions), and the poses
of task-relevant objects read from the simulator. A task is solved when a binary
terminal predicate $\mathcal{G}$ is satisfied.

Instead of mapping observations directly to motor commands, VLCP represents the
policy as a Python function
$\texttt{control}(\textit{obs},\,J) \to \textit{actions}$, where $J \in
\mathbb{R}^{3 \times 7}$ is the end-effector Jacobian and \textit{actions} is a
dict of per-joint position targets consumed by the simulator's PD controller.
The Jacobian is computed analytically at each step and passed to the policy
alongside the observation. This hands velocity-IK to the generated code as a
ready-made primitive.

\subsection{Closed-loop code as a recovery mechanism}
\label{sec:closed-loop}

VLCP's central claim is that recovering from an execution failure means closing
the loop \emph{at the level of the control code}, within the episode. It helps
to separate the two loops the policy runs on. Within a block, the compiled
\texttt{control} function is closed-loop on observations: it gets fresh
\textit{obs} and $J$ every step and can react to numeric state. Its \emph{logic},
though, is frozen. If that logic is wrong, say a mistimed gripper close or a
lateral offset that misses the object, no amount of within-block feedback can fix
it, because the error lives in the code, not in the inputs. So VLCP adds a
second, slower loop across blocks: every $K$ steps the VLM re-observes the scene
and \emph{rewrites the function itself}. That is the loop that makes recovery
possible.

This is exactly what open-loop and monitor-based methods cannot do
(Section~\ref{sec:related}). An open-loop code policy never re-observes, so a
failed grasp rides silently through the rest of the rollout. A monitor sees the
failure but has no code to change: it can only retry the same policy or escalate
to a higher-level planner. VLCP is the one design where the failing artifact, the
control function, is both observed and rewritten mid-episode, at whatever
granularity the failure calls for.

\subsection{Visual closed-loop replanning}

\paragraph{The role of vision.}
Since VLCP is handed privileged object poses, it is fair to ask what the RGB
observations are for. Two things. First, the poses only \emph{seed} the reach.
They locate objects for the initial \texttt{control} block, but they say nothing
about how the attempt then unfolds. The failure signatures the VLM has to read
to recover are diagnosed from the synchronized multi-view RGB at each replan
boundary, not from a coordinate list: a gripper that closed a centimetre short
and lifted empty, an approach angle that clips the object, a grasp that slips, an
object nudged out of place (Section~\ref{sec:exp-recovery}). It is the images
that let the model \emph{see} the miss and rewrite the controller around it.
Second, the privileged poses take object \emph{localization} off the table on
purpose, so the question we actually test, whether closing the loop on the
control code within an episode enables recovery, stays isolated from perception
error rather than tangled up with it. Vision is the channel that drives the
closed-loop correction this paper is about; the poses only bootstrap the reach
it corrects. Swapping them for a learned detector
(Section~\ref{sec:limitations}) is an engineering substitution that leaves this
mechanism intact.

\paragraph{Replan cadence.}
The episode is split into blocks of $K$ steps (default $K{=}50$). At the start of
each block the VLM is queried once to generate a new \texttt{control} function,
which then runs for every remaining step in the block with no further VLM
queries. The policy is thus closed-loop on observations within a block (the same
compiled function gets fresh $\textit{obs}$ and $J$ each step) and closed-loop on
code across blocks (the VLM can rewrite the function at every block boundary).

$K$ trades correction latency against query cost. A smaller $K$ re-observes
more often and catches a failure sooner, but spends more VLM calls per episode;
a larger $K$ is cheaper but lets a failed block run longer before the loop can
intervene. We set $K{=}50$ so that the $500$-step episode horizon divides into
about $10$ replans (Section~\ref{sec:exp-cost}): enough boundaries that a missed
grasp is re-observed within roughly one block of the manipulation timescale,
while keeping the per-episode query count and the cumulative frontier-model
latency (a median $18.8$\,s per replan) within a practical budget. We did not
sweep $K$; a systematic study of the cadence--cost trade-off, including
task-adaptive schedules that replan more densely around contact events, is left
as future work.

\paragraph{Prompt structure.}
Each replan assembles a three-part system prompt followed by a conversation
history and a fresh user turn.

\begin{itemize}

\item \textbf{Block A, static prefix}: robot and environment documentation, the skills API source, and three worked single-phase control examples.

\item \textbf{Block B, learned skill library}: the source of every module currently in the on-disk skill library, sorted by recency and capped at 12 entries.

\item \textbf{Block C, goal pin}: the task identifier and natural-language goal, placed in the system prompt rather than the conversation history so it survives history truncation.

\end{itemize}

The prompt is structured to maximise cache reuse.

The user turn for replan $n$ contains four things: (i) the previous block's
start-of-block RGB images (two views), the generated code, and the $K$ action
dicts it produced, which give the VLM a concrete trace of what was attempted;
(ii) the current synchronized RGB images from both cameras; (iii) a JSON numeric
observation summary (EE pose, joint positions, object poses); and (iv) a prompt
asking for a \texttt{control} function for the next $K$ steps. Images appear only
in the live API turn. The history copy of each turn drops the images and the
prose that refers to them, so no dangling pointers accumulate in the context
window.

\paragraph{History and truncation.}
The conversation history is the sequence of prior user/\allowbreak assistant
replan pairs, kept up to a maximum of $H$ turns (default 20 pairs). Once the
limit is passed, the oldest pairs are dropped from the front. Goal and task
identity live in Block C, so they are never truncated. A cache breakpoint sits
on the last assistant turn, so the prefix through the most recent generated code
is served from cache on the next call.

\subsection{Skill library}

During any replan the VLM may emit one or more labeled helper modules
alongside the \texttt{control} block:

\begin{verbatim}
# skills/move_to_grasp.py
def move_to_grasp(obs, J, target, ...): ...
\end{verbatim}

Each such block is written atomically to a shared on-disk directory
(\texttt{control\_\allowbreak policies/\allowbreak skills/}) and imported right
away. If the import fails, the file is reverted and the error is surfaced to the
VLM on the next replan. Skill files persist across episodes and are shared across
concurrent \texttt{LLMPolicy} instances, coordinated with a process-level lock.

At every replan, Block B is rebuilt: scan the skills directory, sort by
modification time (most recent first), and inject up to 12 module sources
verbatim into the system prompt. So any \texttt{control} function, in this
episode or a later one, can import a skill written earlier.

\section{Experiments}
\label{sec:exp}

\subsection{Setup}
\label{sec:setup}

We evaluate in MuJoCo via RoboVerse (Franka Panda 7-DoF arm with a parallel
gripper). VLCP uses GPT-5.5 as the replanning model, queried every $K{=}50$ steps
as described in Section~\ref{sec:method}. Beyond the LIBERO-Object generalization
study (Section~\ref{sec:exp-vla0}), we run a broad $57$-task sweep across three
scene families: \emph{Object Picks} (the $10$ LIBERO-Object pick tasks),
\emph{Kitchen Scenes} ($30$ tasks), and \emph{Living-Room Scenes} ($17$ tasks).
The latter two come from LIBERO-90 and cover articulated actions, pick-and-place,
distractor disambiguation, elevated and drawer placement, stacking, and
multi-step sequences. Each task runs for a single episode, with the skill
library reset beforehand to keep trials independent, and is pooled within
its family, and we report per-family and overall pooled success rate with Wilson
$95\%$ confidence intervals throughout. The sweep contributes $57$ closed-loop
episodes (one per task) that, with a matched $57$-episode open-loop ($K{=}T$)
run, form the main comparison of Section~\ref{sec:exp-main}. The recovery and
cost analyses (Sections~\ref{sec:exp-recovery} and~\ref{sec:exp-cost}) run
instead over our full logged corpus of closed-loop episodes, a superset of the
sweep that also includes the extra closed-loop rollouts logged during the
instruction-reframing study (Section~\ref{sec:exp-vla0}). Their denominators
($75$ pick-and-place episodes and $81$ closed-loop episodes) are therefore larger
than the $57$-task sweep.

\subsection{Baselines}
\label{sec:exp-baselines}

\textbf{Open-loop VLCP ($K{=}T$)} is our primary controlled baseline: the same
system queried once per episode with no within-episode replan. It is the
\emph{only} comparison that isolates the replanning loop, because it holds the
simulator, control API, privileged poses, task set, model, and episode budget
exactly fixed and varies the single bit this paper is about. Whatever gap it
opens cannot be pinned on any other difference. An external code agent, run on
another benchmark through another API, cannot make that attribution however
strong it is (Section~\ref{sec:exp-main}).

\textbf{VLA-0}~\cite{vla0}, fine-tuned on the full LIBERO benchmark, is a
demonstration-trained \emph{upper reference}, not a matched baseline. We use it
only in the instruction-reframing study, where its in-distribution ceiling makes
the generalization gap easy to read (Section~\ref{sec:exp-vla0}). Comparing raw
success against it would confound demonstration fine-tuning with the closed-loop
question this paper studies.

\textbf{CaP-Agent0}~\cite{capx2026} is the concurrent training-free coding agent
of CaP-X. Like VLCP it generates control code rather than motor tokens and
rewrites it within an episode. We treat it only as an external reference point
for where training-free code agents sit, not as a controlled comparison. It is
evaluated on a different benchmark (LIBERO-PRO) and harness, so we cite its
reported numbers rather than claim a like-for-like comparison
(Section~\ref{sec:exp-main}).

\subsection{Within-episode replanning drives success}
\label{sec:exp-main}

Our central experiment isolates the one design choice this paper is about,
closing the loop on the control \emph{code} within an episode, by comparing VLCP
against its open-loop $K{=}T$ case: the same system with the replan
cadence set to $K{=}T$, so the VLM is queried once at the start of the episode
and the resulting \texttt{control} function, still called and reacting every
step, runs the full horizon with no
re-observation. This $K{=}T$ condition is precisely the Code-as-Policies /
ProgPrompt protocol~\cite{liang2023cap,singh2023progprompt} --- a single
\texttt{control} program written before the arm moves --- realized in our
environment, so the comparison doubles as a matched comparison against the
dominant prior code-as-policy paradigm. The ablation shares VLCP's simulator,
control API, privileged object poses, task set, model, and episode budget, so any
difference traces to the replanning loop alone and not to a change of environment
or interface.

\begin{figure}[t]
\centering
\includegraphics[width=\linewidth]{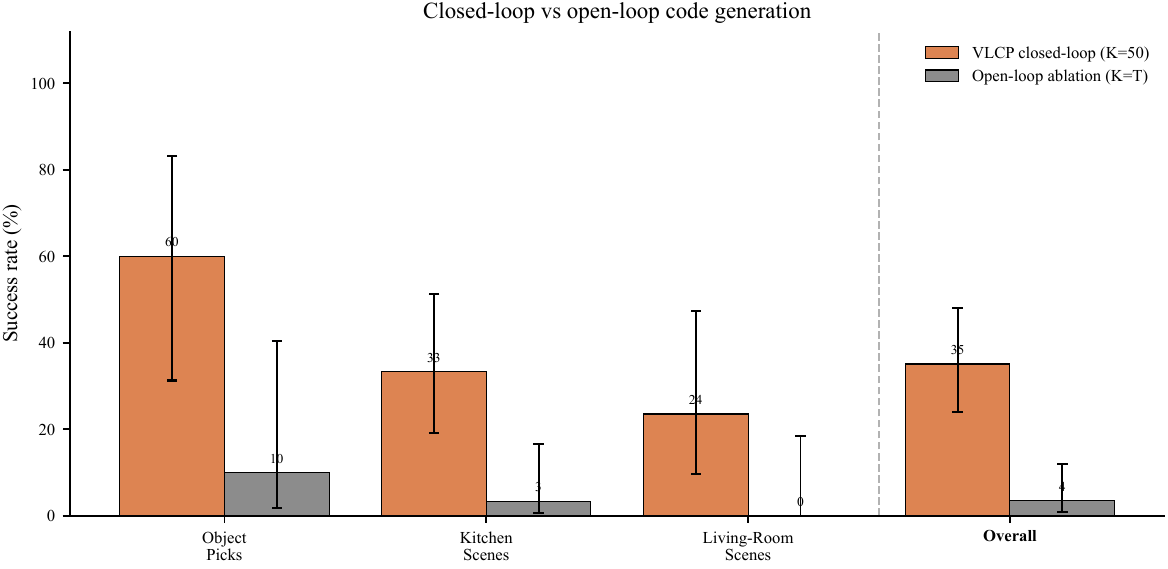}
\caption{\textbf{Within-episode replanning is the dominant factor.} The same
system with the replan cadence set to $K{=}T$ (open-loop, grey) collapses to
near-zero success in every scene family, while VLCP's $K{=}50$ closed-loop
replanning (orange) recovers an order of magnitude more task completions. Bars
are pooled point estimates, whiskers are Wilson $95\%$ CIs, and the dashed
separator marks the pooled Overall group.}
\label{fig:openloop}
\end{figure}

Training-free and with zero demonstrations, closed-loop VLCP reaches $60.0\%$ on
Object Picks, $33.3\%$ on Kitchen Scenes, and $23.5\%$ on Living-Room Scenes,
pooling to $35.1\%$ [$24.0$, $48.1$] overall. Set to $K{=}T$, the same system
collapses to $10.0\%$, $3.3\%$, and $0.0\%$, or $3.5\%$ [$1.0$, $11.9$] pooled.
That is a tenfold gap, with non-overlapping confidence intervals in every scene
family (Figure~\ref{fig:openloop}). The gap is the closed-loop contribution, and
it matches the granularity argument of Section~\ref{sec:closed-loop}. An
open-loop controller cannot revise its code once execution begins. VLCP's fixed
short-horizon replanning catches and repairs deviations before they propagate
through the rest of the episode, which is the recovery behavior we quantify in
Section~\ref{sec:exp-recovery}. Absolute success tracks the available headroom. It
falls from the near-single-step Object Picks to the long-horizon Living-Room
scenes, just as you would expect of a training-free controller that solves the
core manipulation but has less margin on multi-step tasks.

\subsection{Generalization under instruction reframing}
\label{sec:exp-vla0}

VLA-0 is trained on LIBERO-Object using the exact language instructions the
benchmark provides. Does its performance reflect genuine task understanding, or
reliance on those specific instruction strings? To find out, we take 6
LIBERO-Object tasks (\texttt{alphabet\_soup}, \texttt{bbq\_sauce},
\texttt{chocolate\_pudding}, \texttt{cream\_cheese}, \texttt{ketchup},
\texttt{butter}) and build a modified variant of each by reframing the
instruction: the original \textit{``pick up \{object\} and place it in the
basket''} becomes \textit{``put \{object\} in the basket''}. The task and success
condition stay the same. Only the phrasing differs from what VLA-0 saw in
training. On the original in-distribution instructions VLA-0 runs near ceiling
($94$--$98\%$) across LIBERO, far above training-free VLCP. That is the
fine-tuned upper reference (Section~\ref{sec:exp-baselines}) against which the
reframing collapse below should be read, not a matched comparison, since it
mixes demonstration fine-tuning with the loop this paper isolates.

VLA-0 is evaluated over 50 episodes per original task and 10 per modified task,
VLCP over 10 episodes per task variant in both conditions. The 50-episode
original-task numbers for VLA-0 come straight from its reported results. For the
modified tasks, and for VLCP throughout, the episode count is a cost decision.
VLA-0 inference is cheap and local, while every VLCP episode incurs VLM API calls
at each replan, so a matched 50-episode evaluation is prohibitively expensive at
this stage. We report Wilson score confidence intervals throughout, which widen
for the $N{=}10$ cells accordingly, and we read the modified-task comparison as
indicative rather than statistically conclusive at this sample size.

\begin{figure}[t]
\centering
\includegraphics[width=\linewidth]{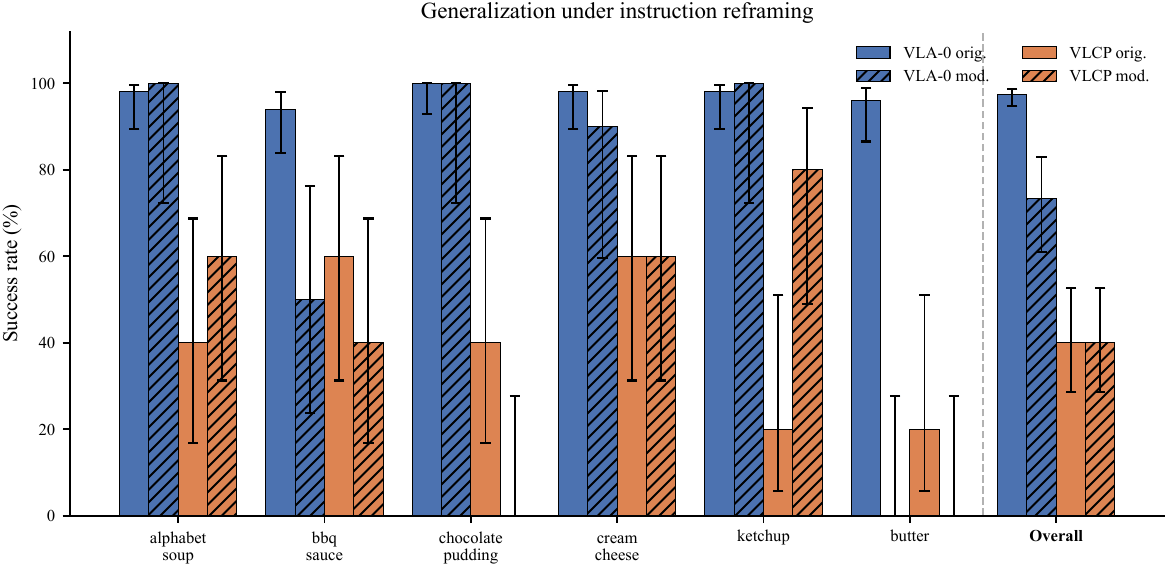}
\caption{Success rate under instruction reframing, per task and pooled overall
(right, dashed separator). Bars are point estimates and whiskers the Wilson 95\%
confidence interval; colour encodes the method (blue: VLA-0, orange: VLCP). Solid
bars: original instruction. Hatched bars: reframed instruction (\textit{``put
\{object\} in the basket''}). VLA-0's response is uneven:
near-ceiling and unchanged on \texttt{chocolate\_pudding} and \texttt{ketchup},
but collapsing outright on \texttt{butter}, for example (non-overlapping
intervals). VLCP (orange) shows no comparable phrasing dependence, and its pooled
success is unchanged under reframing.}
\label{fig:gen-bar}
\end{figure}

Figure~\ref{fig:gen-bar} reports per-task success rate on original and modified
variants for both methods, with the signed change $\Delta = \text{modified} -
\text{original}$ isolating each method's generalization gap. VLA-0's response to
reframing is uneven. It is unchanged or near-ceiling on
\texttt{chocolate\_pudding}, \texttt{ketchup}, and \texttt{alphabet\_soup}, but
it collapses outright on \texttt{butter} (96\% $\to$ 0\%) and drops hard on
\texttt{bbq\_sauce} (96\% $\to$ 50\%), pooling to an overall drop from 97.3\% to
75.0\%. 
% The unevenness is the tell. 
VLA-0's near-ceiling original performance
rests partly on the exact instruction phrasing for at least some tasks, not on
task understanding that survives a surface rewording. VLCP shows no such phrasing
dependence. Its pooled success is unchanged under reframing (40.0\% in both
conditions, $\Delta = 0$), and the per-task differences scatter both ways: gains
on \texttt{ketchup} and \texttt{alphabet\_soup} offset drops on
\texttt{chocolate\_pudding} and \texttt{butter}, with none of the systematic,
task-specific collapse VLA-0 shows on \texttt{butter} and \texttt{bbq\_sauce}. 

% At $N{=}10$ these swings sit inside the wide Wilson intervals and read better as
% sampling noise than as instruction sensitivity, which is what you would expect of
% a controller conditioned on the goal semantics rather than the exact instruction
% string.

\subsection{Recovery behavior}
\label{sec:exp-recovery}

The mechanism VLCP sells is within-episode recovery: a block fails, the next
replan sees the failure and rewrites the controller to fix it. We would rather
not leave that as an anecdote, so we measure how often it happens, directly from
the logged replan traces. On the $75$ closed-loop pick-and-place episodes we
reconstruct the target object's height and the gripper state at every block
boundary from the numeric observation logged with each replan, and we label a
block a \emph{failed grasp} when the hand is at the object (end-effector--object
distance below $0.13$\,m) with the gripper closed but the object still on the
surface. An episode \emph{recovers} when a later block secures and lifts the
object after such a failure. That transition is one an open-loop controller
cannot make, because it never re-observes.

Of the $75$ episodes, $55$ contain at least one failed grasp. In $\mathbf{15}$ of
those ($\mathbf{27.3\%}$) a later replan re-approaches and lifts the object,
recovering a median of $3$ replans after the miss (Table~\ref{tab:recovery}). The
other $40$ never secure the object within the horizon. So recovery is common but
far from guaranteed, and that is precisely the headroom behind the closed-loop
success gain of Section~\ref{sec:exp-main}. The same failed-grasp signature is
terminal for the open-loop variant by construction: its recovery rate is $0$ and
its success collapses to near zero (Figure~\ref{fig:openloop}).

The $40$ non-recovering episodes are where VLCP itself falls short. The loop
keeps re-approaching, but a run of near-misses, a grasp that closes short or
off-centre and sometimes nudges the object, can burn the block budget before a
clean lift lands. Long-horizon tasks make this worse: even a recovered grasp
still leaves placement sub-goals to finish inside the same fixed horizon, which
is why success drops from Object Picks to Living-Room scenes
(Section~\ref{sec:exp-main}). Replanning turns many first-attempt failures into
eventual successes, but it cannot buy more time. A longer horizon or a denser
cadence around contact (Section~\ref{sec:method}) is the lever for the rest.

\begin{figure}[t]
\centering
\setlength{\tabcolsep}{1.5pt}
\begin{tabular}{ccccc}
\includegraphics[width=0.192\linewidth]{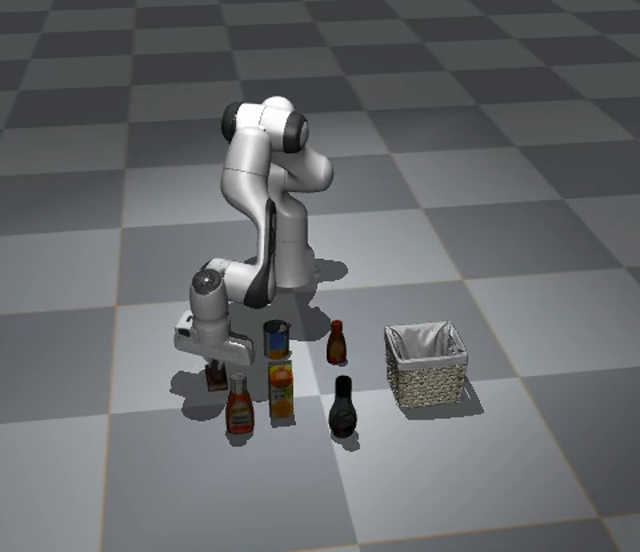} &
\includegraphics[width=0.192\linewidth]{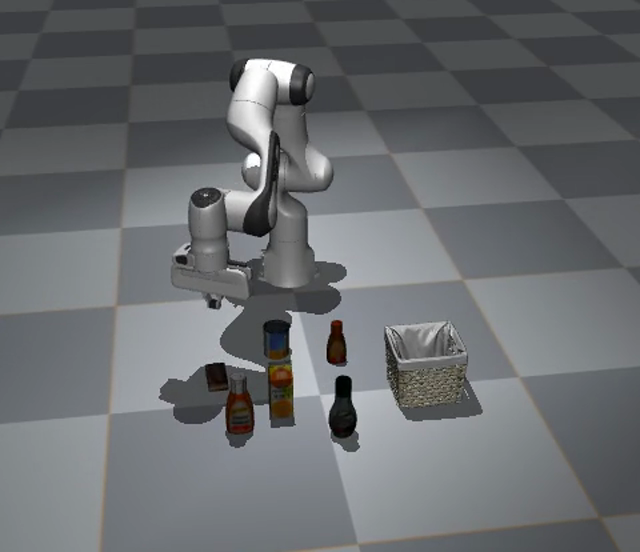} &
\includegraphics[width=0.192\linewidth]{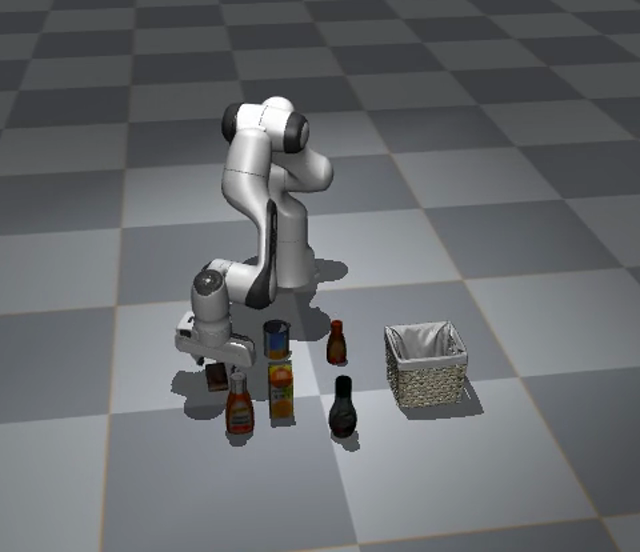} &
\includegraphics[width=0.192\linewidth]{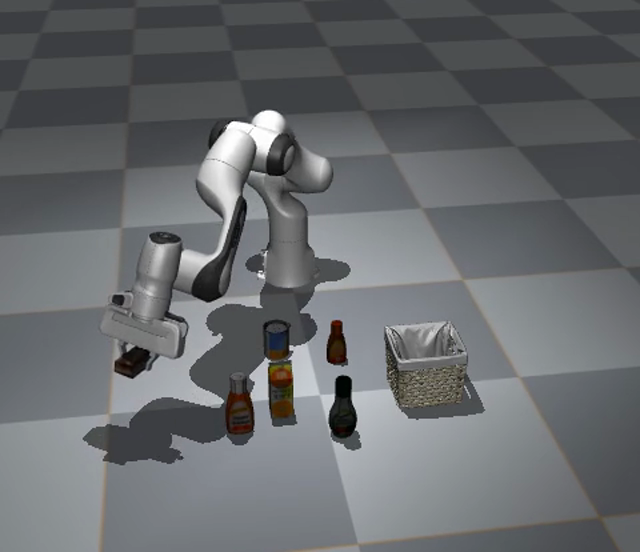} &
\includegraphics[width=0.192\linewidth]{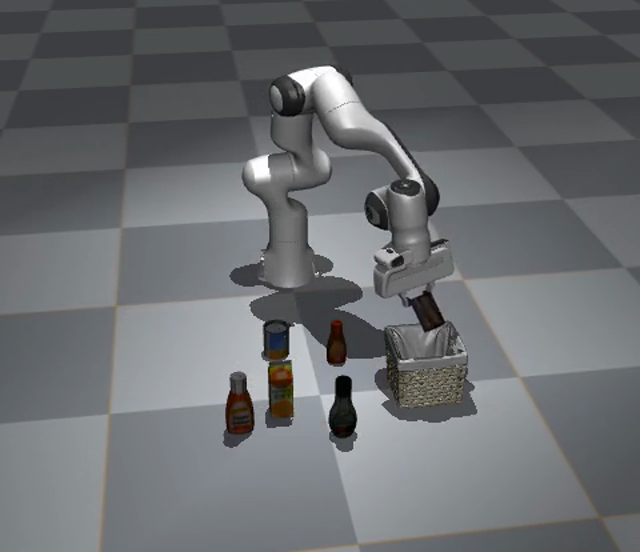} \\
{\footnotesize (a) grasp attempt} &
{\footnotesize (b) grasp fails} &
{\footnotesize (c) replan} &
{\footnotesize (d) grasp succeeds} &
{\footnotesize (e) placed} \\
\end{tabular}
\caption{\textbf{Closed-loop recovery within a single episode}
(\texttt{chocolate\_pudding}, front camera), one instance from the recovery
population of Table~\ref{tab:recovery}. \textbf{(a)} The first control block
reaches for the object and attempts a grasp. \textbf{(b)} The gripper closes
short and lifts \emph{empty}; the object stays on the table. An open-loop
controller would carry this failure to the end of the episode. \textbf{(c)} At
the next replan boundary VLCP re-observes the miss and \emph{rewrites} the
\texttt{control} function, re-approaching the object. \textbf{(d)} The corrected
code grasps and lifts the pudding. \textbf{(e)} It goes into the basket and the
task succeeds. Recovery works because the loop closes on the control code itself,
not just on the observations fed to a fixed policy.}
\label{fig:recovery}
\end{figure}

Reading the recovery episodes confirms the transitions are model-driven, not
physical accidents. At the corrective block the model diagnoses the miss in the
code it writes. In one episode it resolves to \emph{``recenter lower and regrasp
the pudding after the last miss \dots\ the pudding [is] still on the floor \dots\
the previous block closed before the hand got low enough''}, then descends and
closes again, and the next block's code reports that \emph{``the previous lift
succeeded.''} The single-episode trace in Figure~\ref{fig:recovery} is one
instance from this population. Failed VLA-0 rollouts look different under the same
scrutiny. Once a failure starts, the policy typically stalls or repeats the same
error, grasping just behind the object or opening the gripper behind the
receptacle, with no way to rewrite the behavior that caused it.

\begin{table}[t]
\centering
\caption{Within-episode recovery on the $75$ closed-loop pick-and-place episodes,
computed directly from the logged per-block object height and gripper state
(\texttt{paper/analysis/recovery\_rate.py}). A \emph{failed grasp} is a block with
the hand at the object and the gripper closed but the object not lifted; an
episode \emph{recovers} if a later block secures and lifts it. Open-loop ($K{=}T$)
recovery is $0$ by construction, since it never re-observes.}
\label{tab:recovery}
\begin{tabular}{lc}
\toprule
Quantity & Value \\
\midrule
Pick-and-place episodes (closed-loop)      & 75 \\
\quad with $\geq 1$ failed grasp           & 55 \\
\quad that recover and lift the object     & 15 \\
Recovery rate (recovered\,/\,failed)       & \textbf{27.3\%} \\
Median replans to recover                  & 3 \\
Open-loop ($K{=}T$) recovery rate          & 0\% (by construction) \\
\bottomrule
\end{tabular}
\end{table}

\subsection{Cost, latency, and prompt caching}
\label{sec:exp-cost}

Since VLCP re-queries the VLM at every block boundary, whether the method is
practical comes down to whether that query is cheap. We report operating cost
measured directly from the logged replan traces of our closed-loop runs ($81$
episodes, $\approx\!721$ replan calls), summarized in Table~\ref{tab:cost}.

Three properties keep the per-replan cost in check. The first is the prompt
layout. A long static prefix (Block~A documentation and API, Block~B skill
library, Block~C goal pin; $\approx\!8$k tokens) is followed by a short fresh
suffix, so on every replan after the first a median of $\mathbf{84\%}$ of the
input tokens fall inside the cached prefix and are billed and processed at the
cache-read rate instead of recomputed. The second is compact output: the model
emits a median of $\approx\!350$ tokens per replan (mean $421$), so decode
latency stays low despite the large context. The third is call count. An episode
takes only a handful of queries, a median of $10$ replans (one per $K{=}50$-step
block over the $500$-step horizon), rather than one query per control step as in
a raw VLA. End-to-end latency for a single replan is a median of $18.8$\,s (mean
$25.9$\,s), dominated by frontier-model server-side generation. That sets the
practical floor on VLCP's control rate, the concrete cost behind the latency
limitation of Section~\ref{sec:limitations}.

Code validity, finally, is not a bottleneck. \textbf{All $778$ generated
\texttt{control} blocks across every episode parsed and compiled successfully}
($100\%$), so the compile-failure safety fallback of Section~\ref{sec:method}
fired only on transient runtime exceptions, never on malformed generations. The
upshot: prompt caching makes within-episode replanning cheap enough that the
order-of-magnitude success gain over the open-loop variant
(Section~\ref{sec:exp-main}) costs roughly ten cache-warm VLM calls per episode.

\begin{table}[t]
\centering
\caption{Measured operating cost of VLCP, aggregated over the closed-loop runs
($81$ episodes, $\approx\!721$ replan calls total). Latency is wall-clock per
replan call, recovered from the logged request/response timestamps. The
cached-prefix fraction is the share of each replan's input tokens lying in the
static, cache-eligible prompt prefix, measured over the $640$ post-first calls
that can hit cache.}
\label{tab:cost}
\begin{tabular}{lcc}
\toprule
Quantity & Median & Mean \\
\midrule
Replan (VLM) calls per episode      & 10    & 8.9  \\
Static prompt prefix (tokens)       & 7.9k  & 8.2k \\
Input tokens per replan             & 43k   & 43k  \\
Cached-prefix fraction per replan (\%) & 83.8 & 79.4 \\
Output tokens per replan            & 349   & 421  \\
Latency per replan (s)              & 18.8  & 25.9 \\
Code validity (parsed \& compiled)  & \multicolumn{2}{c}{$778/778 = 100\%$} \\
\bottomrule
\end{tabular}
\end{table}

\section{Conclusion}
\label{sec:conclusion}

We introduced VLCP, a training-free framework that closes the manipulation loop
at the level of the control code itself. A frozen VLM is re-queried every $K$
steps and \emph{rewrites} the control function from live multi-view RGB,
proprioceptive state, and an explicit state delta, so a failure is caught and
corrected while the episode is still unfolding. That is a different place to
close the loop than prior work chose. VLCP does not retry a fixed policy or
reselect a subtask. It edits the failing artifact directly, and the loop, not any
single generated program, is the contribution.

We think the evidence is hard to pin on anything else. Against its own open-loop
degenerate case ($K{=}T$), an ablation that holds the simulator, control API,
privileged poses, task set, model, and episode budget fixed and varies
\emph{only} within-episode replanning, closing the loop lifts pooled success from
$3.5\%$ to $35.1\%$: a tenfold gain, non-overlapping confidence intervals in
every scene family, across a $57$-task sweep. And we tie that gain to a concrete,
measured mechanism rather than leaving it a black box. $27.3\%$ of failed grasps
are re-approached and lifted by a later replan, a transition the open-loop
variant cannot make by construction, and the corrective code names the miss it is
fixing. The loop is affordable as well as effective. Prompt caching serves a
median $84\%$ of input tokens from cache and an episode costs only $\sim$10
compact queries, so the order-of-magnitude gain runs on roughly ten cache-warm
calls.

\paragraph{Limitations.}
\label{sec:limitations}
VLCP relies on simulator-provided object poses and has only been run in
simulation, so its robustness to perception noise and real-robot dynamics is
untested. We factor localization out on purpose, to isolate the closed-loop
question, and swapping the poses for a learned detector is an engineering
substitution that leaves the replanning mechanism intact. Replanning also brings
VLM latency and API cost. Each replan takes a median of $18.8$\,s of
frontier-model inference (Section~\ref{sec:exp-cost}), so even with prompt caching
bounding the token cost, the method suits relatively slow manipulation rather than
high-frequency reactive control. That same API cost also bounds our evaluation
budget. Because every episode incurs closed-source VLM calls at each replan, the
main sweep runs a single episode per task (Section~\ref{sec:setup}), and we lean
on per-family pooling and Wilson confidence intervals rather than multi-seed
repetition. A multi-seed evaluation that quantifies per-task variance is the
right next step, and moving to an open-weight but capable VLM would make it
affordable. Performance also depends on a strong frozen VLM.
The smaller open models we tried (Qwen3-4B) failed to produce reliable control
code or effective replanning, which limits accessibility and raises deployment
cost. We
read all of these as constraints on the present instantiation, not on the
underlying principle of closing the loop on control code.

\paragraph{Outlook.}
The direction we find most compelling is the cross-episode skill library. Because
successful control blocks are already saved and re-injected into later prompts,
the repertoire grows with experience \emph{without retraining}, which gives VLCP
a natural substrate for \emph{continual learning}: later episodes reuse and
compose behaviors written earlier, and the same closed-loop machinery that
recovers within an episode starts to accumulate competence across them. Scaling
that library, and measuring how the accumulated repertoire transfers across
increasingly diverse tasks and onto real-robot perception and dynamics, is where
we expect a training-free, self-editing code policy to compound. More broadly,
VLCP points to a path to capable robot policies that skips demonstration
collection entirely, by keeping the model in the representation it is already most
fluent in. We hope the code-level closed loop proves useful well beyond the tasks
studied here.

\section{Acknowledgement}
\label{sec:acknowledgement}
This work was conducted as part of the Algoverse AI Research program.

% \section*{Reproducibility Statement}
% \label{sec:repro}
%
% Code: \texttt{llm\_policy.py} / \texttt{llm\_policy\_libero.py} (the closed-loop
% replanning runner), plus skills, prompts, exact task ids, model strings, and
% seeds. Every replan is logged to disk (redacted request, raw response, per-view
% images, and generated code per block). The cost, latency, cache, and
% code-validity figures of Section~\ref{sec:exp-cost} are computed directly from
% these per-block logs, and the full request/response payloads with token and
% cache-read counts are also captured as Phoenix traces. Appendix~\ref{sec:app-prompt}
% reproduces the prompt assets verbatim (the output-format contract and one of the
% worked \texttt{control} examples shipped in Block~A), and
% Appendix~\ref{sec:app-generated} shows a full \texttt{control}-block skill the
% frozen VLM wrote unedited during a recovery episode.

% ---- Bibliography ----
\bibliographystyle{splncs04}
\bibliography{main}

\appendix

\section{Prompt assets}
\label{sec:app-prompt}

The system prompt is assembled from the three cache-structured blocks of
Section~\ref{sec:method} (A~documentation/API/examples, B~skill library, C~goal
pin). We reproduce here the two pieces of Block~A that most directly determine
what the VLM emits: the strict output-format contract, and one of the three
worked \texttt{control} examples the prompt ships with.

\paragraph{Output-format contract (verbatim).}
The preamble instructs the frozen VLM to respond only with header-tagged code
blocks, defining exactly one \texttt{control} function and any number of reusable
skill modules:

\footnotesize
\begin{verbatim}
Output format (strict):
- Respond with ONE OR MORE fenced ```python``` blocks and nothing else
  outside those blocks. No prose before, between, or after.
- Each block MUST be preceded on its own immediately-prior line by a
  header comment naming its destination:
    `# control` — exactly one such block per response; it must define
    `def control(obs, J) -> list[dict]` with the same contract as the
    worked examples below.
    `# skills/<snake_case_name>.py` — zero or more reusable helper
    modules. Each is persisted to disk and importable as
    `from skills.<name> import <fn>` on this and every subsequent
    replan, including by other concurrent runs in the same library.
- You MAY keep module-level state (e.g. a `_STATE` dict with a phase
  field) inside the `# control` block — the function is called K times
  in a row with fresh `obs`/`J` before you are queried again. Do NOT
  keep mutable state inside skill files.
- Each replan only needs to handle the *next* phase of the task
  (~K closed-loop steps), not the full multi-phase plan.
- Inspect all provided images carefully. The synchronized camera views
  often contain the most reliable information about the real state of
  the world, including whether a grasp or push actually succeeded, and
  what changed since the previous replan.
\end{verbatim}
\normalsize

\paragraph{Worked \texttt{control} example (verbatim).}
One of the three single-phase examples in Block~A, demonstrating the
\texttt{control(obs,\,J)} contract, the observation-access idiom, and explicit
gripper use through the \texttt{move\_to\_target} IK primitive:

\footnotesize
\begin{verbatim}
"""Single-phase example: close the gripper on the milk carton.

Scope: ONLY the grasp/settle phase. Assume a prior block has already
descended the hand to the grasp height with the fingers open. This
`control` function holds the end-effector at its current position while
commanding both finger joints closed, so the pads can clamp the milk
before a later block lifts it.
"""

import numpy as np
from skills.move_to_target import move_to_target

GRIPPER_CLOSE = 0.0

def control(obs: "TensorState", J: np.ndarray) -> list[dict]:
    arm_q = obs.robots["franka"].joint_pos[0].numpy()[2:9]
    ee_pos = obs.robots["franka"].body_state[0, 0, :3].numpy()

    # Hold position while closing. Avoid null-space motion here so the
    # hand does not drift off the carton as the fingers clamp.
    return move_to_target(
        arm_q, J, ee_pos, ee_pos, GRIPPER_CLOSE,
        lam=0.02, max_step=0.05, null_gain=0.0, dq_cap=0.05,
    )
\end{verbatim}
\normalsize

\section{Generated-code sample}
\label{sec:app-generated}

During a chocolate-pudding pick-and-place episode --- the same task family as
the recovery trace in Figure~\ref{fig:recovery} --- the frozen VLM emitted the
skill module below, unedited. It was written to
\texttt{skills/\allowbreak chocolate\_\allowbreak pudding\_\allowbreak
basket\_\allowbreak targets.py} during a replan and re-injected verbatim into every later replan's Block~B, where
the episode's \texttt{control} function imports it. It illustrates the format the
VLM was asked to follow: a module docstring stating scope and assumptions, full
type hints, and per-function Args/Returns docstrings.

\footnotesize
\begin{verbatim}
"""Target helpers for picking chocolate pudding and placing it in a basket.

Scope:
    This module computes simple world-frame panda_hand position targets
    for top-down manipulation of the small chocolate pudding object and
    placement into a nearby basket in LIBERO grocery-style scenes.

Assumptions:
    - The chocolate pudding root position is close to the center of the
      object footprint in XY.
    - A top-down parallel-jaw grasp is used, with the panda_hand frame
      roughly 10 cm above the fingertip pads.
    - Returned targets are panda_hand positions in world coordinates,
      not object poses or fingertip positions.
"""

from typing import Tuple
import numpy as np


def chocolate_pudding_pick_targets(
    pudding_pos: np.ndarray,
    hover_z: float = 0.235,
    grasp_z: float = 0.118,
    lift_z: float = 0.255,
    xy_offset: np.ndarray | None = None,
) -> Tuple[np.ndarray, np.ndarray, np.ndarray]:
    """Compute hover, low grasp, and lift targets for chocolate pudding.

    Args:
        pudding_pos: (3,) world position of the chocolate pudding root.
        hover_z: World-frame panda_hand height for the approach hover.
        grasp_z: World-frame panda_hand height for the low grasp.
        lift_z: World-frame panda_hand height after closing.
        xy_offset: Optional (2,) XY offset from the pudding root to the
            desired gripper center. None uses zero offset.

    Returns:
        Tuple (hover_target, grasp_target, lift_target) of (3,) arrays
        in world coordinates.
    """
    offset = np.zeros(2) if xy_offset is None else xy_offset.astype(float)
    xy = pudding_pos[:2].astype(float) + offset
    hover_target = np.array([xy[0], xy[1], float(hover_z)])
    grasp_target = np.array([xy[0], xy[1], float(grasp_z)])
    lift_target = np.array([xy[0], xy[1], float(lift_z)])
    return hover_target, grasp_target, lift_target


def chocolate_pudding_basket_place_target(
    basket_pos: np.ndarray,
    hand_z: float = 0.175,
    xy_offset: np.ndarray | None = None,
) -> np.ndarray:
    """Compute a panda_hand target for placing pudding in the basket.

    Args:
        basket_pos: (3,) world position of the basket root.
        hand_z: World-frame panda_hand height for lowering the held
            pudding into the basket opening before release.
        xy_offset: Optional (2,) XY offset from the basket root to the
            desired placement point. None uses the basket center.

    Returns:
        A (3,) world-frame panda_hand target inside the basket opening.
    """
    offset = np.zeros(2) if xy_offset is None else xy_offset.astype(float)
    xy = basket_pos[:2].astype(float) + offset
    return np.array([xy[0], xy[1], float(hand_z)])
\end{verbatim}
\normalsize

\end{document}